\documentclass{article} 
\usepackage{colm2024_conference}

\usepackage{microtype}
\usepackage{hyperref}
\usepackage{url}
\usepackage{booktabs}
\usepackage{graphicx}
\usepackage{kotex}
\usepackage{multirow}
\usepackage{authblk}

\colmfinalcopy 
\begin{document}

\title{Enhancing Clinical Efficiency through LLM: \\Discharge Note Generation for Cardiac Patients
}

\author[1]{HyoJe Jung}
\author[1]{Yunha Kim}
\author[1]{Heejung Choi}
\author[2]{Hyeram Seo}
\author[2]{Minkyoung Kim}
\author[2]{JiYe Han}
\author[1]{Gaeun Kee}
\author[1]{Seohyun Park}
\author[1]{Soyoung Ko}
\author[1]{Byeolhee Kim}
\author[3]{Suyeon Kim}
\author[4]{Tae Joon Jun}
\author[5]{Young-Hak Kim}

\begin{center}
    \tiny
    \affil[1]{Department of Medical Science, Asan Medical Institute of Convergence Science and Technology, Asan Medical Center, University of Ulsan College of Medicine, 88, Olympicro 43gil, Songpagu, 05505, Seoul, Republic of Korea}
    \affil[2]{Department of Information Medicine, Asan Medical Center, 88, Olympicro 43gil, Songpagu, 05505, Seoul, Republic of Korea}
    \affil[3]{INMED DATA, 88, Olympicro 43gil, Songpagu, 05505, Seoul, Republic of Korea}
    \affil[4]{Big Data Research Center, Asan Institute for Life Sciences, Asan Medical Center, 88, Olympicro 43gil, Songpagu, 05505, Seoul, Republic of Korea}
    \affil[5]{Division of Cardiology, Department of Information Medicine, Asan Medical Center, University of Ulsan College of Medicine,  88, Olympicro 43gil, Songpagu, 05505, Seoul, Republic of Korea}
\end{center}

\maketitle

\begin{abstract}
Medical documentation, including discharge notes, is crucial for ensuring patient care quality, continuity, and effective medical communication. 
However, the manual creation of these documents is not only time-consuming but also prone to inconsistencies and potential errors. 
The automation of this documentation process using artificial intelligence (AI) represents a promising area of innovation in healthcare.

This study directly addresses the inefficiencies and inaccuracies in creating discharge notes manually, particularly for cardiac patients, by employing AI techniques, specifically large language model (LLM). Utilizing a substantial dataset from a cardiology center, encompassing wide-ranging medical records and physician assessments, our research evaluates the capability of LLM to enhance the documentation process.

Among the various models assessed, Mistral-7B distinguished itself by accurately generating discharge notes that significantly improve both documentation efficiency and the continuity of care for patients. These notes underwent rigorous qualitative evaluation by medical expert, receiving high marks for their clinical relevance, completeness, readability, and contribution to informed decision-making and care planning. Coupled with quantitative analyses, these results confirm Mistral-7B’s efficacy in distilling complex medical information into concise, coherent summaries.

Overall, our findings illuminate the considerable promise of specialized LLM, such as Mistral-7B, in refining healthcare documentation workflows and advancing patient care. This study lays the groundwork for further integrating advanced AI technologies in healthcare, demonstrating their potential to revolutionize patient documentation and support better care outcomes.
\end{abstract}

\section{Introduction}

Discharge notes are essential in the healthcare sector, serving as a comprehensive summary of a patient's hospital stay, including diagnosis, treatments, and follow-up care recommendations. 
These documents are crucial for ensuring smooth transitions between care settings, enhancing communication among healthcare providers, and supporting effective patient management after hospitalization. 
Furthermore, they play a significant role in reducing readmission rates by facilitating the proper management of ongoing care plans, thus maintaining the quality and safety of patient care.

With the advancement of AI, especially natural language processing (NLP) algorithms, there has been a growing interest in their application across various domains, including healthcare. 
NLP technologies have shown capability in automating tasks that require understanding and generating human language, transforming the way we interact with data and technology. 
In healthcare, these applications range from automating clinical documentation to enhancing patient interaction with care providers through conversational agents.

LLM has found significant applications in the medical field, including creating clinical notes, interpreting lab results, and anonymizing patient data \citep{llm_medical-field}. 
Their ability to generate human-like text suggests to increase both efficiency and accuracy in medical documentation, potentially reducing the administrative workload for healthcare professionals and allowing them more time for patient care. 
However, implementing LLM in healthcare faces challenges, notably the linguistic diversity and medical jargon prevalent in multi-cultural hospital settings. 
This complexity demands LLM to comprehend and interpret medical terminology across different languages and understand the nuances of clinical communication.

Our study underscores the pivotal role of deploying specialized LLM, particularly Mistral-7B, in automating the creation of discharge notes within cardiology, utilizing actual hospital patient data. By expertly navigating the complexities of genuine patient records, Mistral-7B significantly elevates documentation efficiency and continuity. Endorsed by cardiology professional for its clinical relevance and utility, Mistral-7B's ability to streamline healthcare documentation with real-world data showcases a promising avenue towards integrating AI-driven tools in enhancing patient care and optimizing medical record accuracy. This advancement marks a crucial step forward in leveraging cutting-edge AI to refine healthcare documentation practices directly from the front lines of patient care.

\section{Related works}
\label{gen_inst}

The application of NLP in the medical field has been experiencing a steady increase. This trend is reflected in numerous studies and projects that leverage NLP techniques to extract valuable insights from medical texts, enhance patient care, and facilitate medical research. As such, our work contributes to this growing body of research, further demonstrating the potential of NLP in healthcare.

\textbf{Clinical Notes Generation} The generation of clinical notes, particularly radiology reports \citep{clinical_notes}, represents a critical intersection between NLP and clinical practices. Traditional methods of generating radiology reports have been both time-consuming and tedious for radiologists, prompting the exploration of automated systems. Recent advances have seen the development of multi-modal approaches that combine images, disease labels, and textual data to generate comprehensive radiology reports. 

\textbf{Patient Summary Generation} Generating patient summaries \citep{discharge_summary} is an area where NLP is making strides, addressing the need for efficient tools to concisely capture crucial patient information. 
Recent work proposes an extractive-abstractive summarization pipeline that extracts key sentences from clinical notes and abstracts them into coherent summaries. 
This approach aids in handling large text volumes while ensuring faithfulness and traceability to original documents. 
NLP applications in clinical note and patient summary generation can enhance accuracy, efficiency, and comprehensiveness of medical documentation, supporting better patient outcomes and alleviating healthcare professionals' workload.

Unlike previous study, we directly generate comprehensive discharge records, including chief complaints, medical history, hospital course, discharge status, and follow-up instructions, using large language models. 
Additionally, we have developed a model specialized in cardiac-related medical terminology and document writing, utilizing a large dataset from a cardiology department of actual hospital.

\textbf{Large Language Model (LLM)} Recent years have witnessed the rise of powerful LLM that have revolutionized various natural language processing tasks, including text generation, summarization, and question answering. These models, trained on vast amounts of textual data, have demonstrated remarkable capabilities in understanding and generating human-like text. 

One of the pioneering open source models in this domain is Llama \citep{llama}, developed by Meta AI. Llama is a family of models ranging from billions to trillions of parameters, trained on a vast corpus of online data. These models have exhibited impressive performance on a wide range of tasks, including open-ended generation, question answering, and code generation. Llama's scalability and adaptability have made it a popular choice among researchers and developers.

Another notable open soruce LLM is Mistral \citep{mistral}, developed by Mistral AI. This model has demonstrated remarkable performance across numerous benchmarks, outperforming all other models of the same size on many tasks. Notably, it has outperformed the larger-sized model of Llama across various benchmarks. With its impressive performance across various benchmarks, innovative architectural design, and specialization for conversational tasks, Mistral has emerged as a powerful and versatile language model, poised to make significant contributions in natural language processing and its applications.

\section{Dataset}
\label{headings}

\textbf{\subsection{Ethical approval}}

This study's protocols received approval from the Asan Medical Center Institutional Review Board (IRB No.2023-1001), aligning with the principles outlined in the 2008 Declaration of Helsinki. Moreover, the need for informed consent was waived due to the utilization of an anonymous, de-identified database for research purposes.

\textbf{\subsection{Data}}
This study utilizes a comprehensive dataset derived from the Cardiology Department of Asan Medical Center, focusing on patients admitted for care. 
The data was collected through the Asan Biomedical Research Environment (ABLE) system, which ensures a high standard of data integrity and relevance for clinical research \citep{ABLE}. 

The dataset encompasses patient records spanning from September 2018 to December 2021, providing a broad temporal snapshot of patient care within the institution.  
For the sake of computational resource and time efficiency, our final dataset, comprising 4,588 unique patient records, was established, with each record resulting in less than 2,048 tokens upon tokenization. These records were then segregated into training, validation, and testing sets for effective model development and evaluation.
Specifically, the dataset is divided into 4,077 records for training, 122 records for validation, and 459 records for testing purposes. 
This division allows for comprehensive training of the models while ensuring robust validation and testing to evaluate the performance accurately.

In this research, the Progress Notes documenting the detailed course of the patient's treatment were utilized as the input data for our model, while the Discharge Notes reflecting the patient's status at the point of discharge served as the target or label data.
The Progress Notes includes fields such as RECORD DATE, PROBLEM LIST, SUBJECTIVE, OBJECTIVE, ASSESSMENT, GOAL, PLAN, and COMMENT, offering a detailed account of the patient's clinical status and treatment plan during their stay.  
Meanwhile, the Discharge Notes are composed of sections detailing the CHIEF COMPLAINT, OPERATION AND PROCEDURE, HOSPITAL COURSE, CONDITION AT DISCHARGE, and TYPE OF DISCHARGE. 
This structure ensures a comprehensive overview of the patient's hospital journey, from admission to discharge, providing a valuable foundation for automating the generation of discharge notes using LLM.

\section{Methods}
\label{headings}

\begin{figure}[h]
\begin{center}
\includegraphics[width=\textwidth]{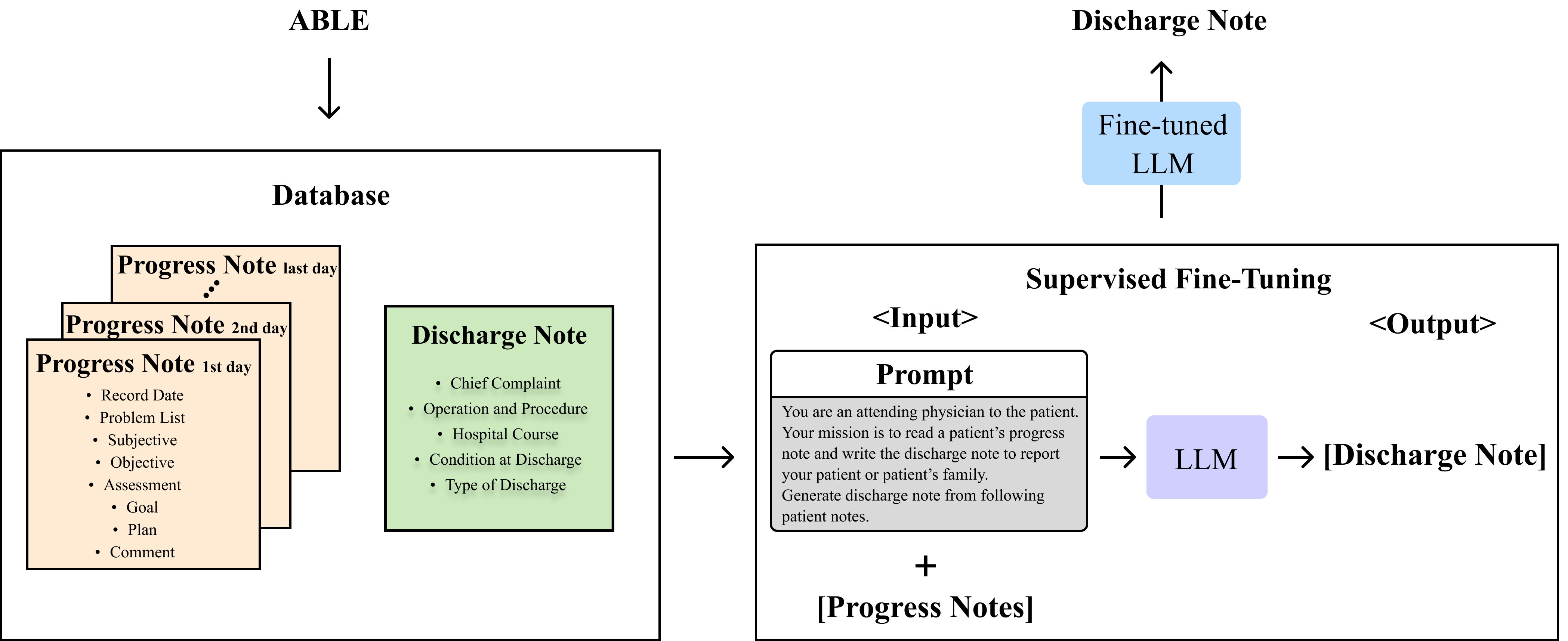}

\end{center}
\caption{A pipeline for generating a discharge note based on a progress note using fine-tuned LLM.}
\end{figure}

\textbf{\subsection{Models}}

In the advancement of our study, we embarked on employing a diverse array of models to automate the generation of discharge notes from detailed patient records. 
Utilizing the supervised fine-tuning (SFT) technique, we tapped into the extensive capabilities of several pre-trained LLM to refine their performance towards our specialized medical documentation task. 
Specifically, we utilized models such as TinyLlama-1.1B \citep{tinyllama}, Llama2-7B \citep{llama}, Mistral-7B \citep{mistral}, BioMistral-7B \citep{biomistral}, and Meditron-7B \citep{meditron}, SOLAR-10.7B \citep{solar}, each selected for their unique strengths and potential in handling complex language tasks pertinent to the medical domain. 
We also utilized the cutting-edge tool, the Unsloth library \citep{unsloth}, to facilitate the fine-tuning process of language models, optimize VRAM usage, and significantly accelerate training speed, thereby enhancing the efficiency of our model's learning process.

\textbf{\subsection{Parameter Efficient Fine Tuning (PEFT)}}

A significant technique incorporated into our model fine-tuning approach is Quantized Low Rank Adaption (QLoRA) \citep{qlora}, a method under of Parameter Efficient Fine Tuning (PEFT) \citep{peft}.
This methodology is optimally designed to overcome limitations arising from limited computational resources while maintaining performance levels.

PEFT emerges as a strategic fine-tuning method to counterbalance the substantial computational demand and memory usage associated with leveraging Transformer-based LLM, particularly those encompassing billions of parameters. 
The essence of PEFT lies in its ability to fine-tune PLM with a minimal subset of parameters, thereby maintaining their performance on specific tasks.
This approach significantly mitigates the challenges inherent in deploying these sophisticated models, especially in settings where computational resources are at a premium. 

The adaptation of PEFT methodologies enables the efficient utilization of LLM for generating medically accurate and contextually precise discharge notes, aligning with our objectives of streamlining healthcare documentation processes.

\textbf{\subsubsection{LoRA Parameters}}

The LoRA fine-tuning process employed the following hyperparameters:

\begin{itemize}
    \item r : 8
    \item lora alpha : 16
    \item lora dropout : 0
    \item target module : q, k, v, o, gate, up, down
\end{itemize}

\textbf{\subsection{Supervised Fine Tuning (SFT)}}

Supervised Fine Tuning (SFT) is a crucial component of our methodology, aiming to adapt pre-trained language model (PLM) for the specialized task of generating discharge notes from patient medical records. Unlike the broader, unsupervised training approaches traditionally associated with LLM, SFT leverages labeled datasets, specifically curated to mirror the task at hand. The fundamental distinction of SFT lies in its utilization of previously validated responses, ensuring that the fine-tuning process is informed by data that has been reliably categorized according to the desired outputs.

Through supervised fine-tuning, our LLMs are intricately molded to recognize and replicate the complex patterns and nuances inherent in the domain-specific data. The adaptation of the model's parameters to the particular distribution of the labeled data and the explicit requirements of generating medically coherent and accurate discharge notes ensures that the model becomes highly proficient in this task. Consequently, SFT empowers the pre-trained LLM to transition from their general language understanding capabilities to specialized expertise required for the precise and effective generation of discharge notes, central to enhancing patient care documentation in healthcare settings. This process not only improves the models' performance on the target task but also enriches their contextual understanding relevant to the medical documentation domain, thereby facilitating the creation of highly reliable discharge notes informed by validated data.

\textbf{\subsubsection{SFT Parameters}}

The supervised fine-tuning process employed the following hyperparameters:
\begin{itemize}
    \item batch size : 8
    \item gradient accumulation step : 4
    \item warmup step : 10
    \item optimizer : adamw\_8bit
\end{itemize}
\textbf{\subsection{Evaluation}}

In our study, we employed both quantitative and qualitative evaluations to assess the model performance. For the quantitative evaluation, we utilized several common metrics. 
\begin{itemize}
    \item Recall-Oriented Understudy for Gisting Evaluation (ROUGE) : ROUGE is a set of metrics used for evaluating automatic summarization and machine translation. It compares the output with reference summaries to measure the quality, focusing on the overlap of n-grams, words, or bytes.
    \item Bilingual Evaluation Understudy (BLEU) : BLEU is a metric for evaluating a generated sentence to a reference sentence. It calculates the precision of n-grams in the generated sentence that also appear in the reference sentence, offering a quantitative measure for translation quality.
    \item BERT Score: BERT Score is a metric for evaluating text generation tasks by calculating the similarity of token embeddings between the generated and reference texts. It leverages the BERT model's ability to capture complex semantic representations, providing a more nuanced evaluation.
    \item Perplexity: Perplexity is a measurement of how well a probability model predicts a sample. In language modeling, it quantifies the uncertainty of predicting the next token in a sequence, with lower perplexity indicating better prediction performance.
\end{itemize}

These measures allowed us to objectively evaluate the performance of our model in terms of various aspects such as precision, recall, semantic coherence, and language fluency.

On the other hand, for the qualitative evaluation, we obtained expert judgement from a professional in Cardiology. This allowed us to incorporate a professional's assessment into our evaluation process, providing a more comprehensive and practical perspective on the usability and accuracy of our generated content. The evaluation was conducted based on following five criteria.

\begin{itemize}
    \item Accuracy : Evaluates how accurately the generated discharge note reflects the patient's actual medical condition, treatment process, and recommended follow-up actions.
    \item Completeness : Assesses whether all important medical information (diagnosis, treatment methods, observations of improvement or deterioration, discharge criteria, and follow-up actions) is included in the discharge note.
    \item Readability and Comprehensibility : Evaluates whether the generated document is easy to understand and clear for the target audience (patients, guardians, other medical professionals, etc.).
    \item Consistency : Assesses whether the discharge notes generated across various patient records maintain a consistent format and quality.
    \item Utility : Evaluate whether the documentation generated is specifically helpful in making clinical decisions and contributes to the development of follow-up care plans.
\end{itemize}

\textbf{\subsection{Environment}}
The experiments were conducted on an Ubuntu 22.04 LTS system with NVIDIA RTX 3090 GPU. The following software versions were used:
\begin{itemize}
    \item Python 3.10.12
    \item Transformers 4.38.2
    \item torch 2.1.1+cu118
    \item TRL 0.7.7
    \item CUDA 11.8
\end{itemize}
\section{Results}
\textbf{\subsection{Quantitative result}}


Table 1 presents the evaluation results of various fine-tuned language models on the task of generating discharge notes for cardiac patients, using multiple quantitative metrics. 

While no single model outperformed across all metrics, the quantitative results collectively highlight the potential of fine-tuned language models to generate accurate, coherent, and clinically relevant discharge notes from patient medical records.

\textbf{\subsection{Qualitative result}}

\begin{table}[]
\centering
\resizebox{\textwidth}{!}{%
\begin{tabular}{ccccccc}
\hline
\multirow{2}{*}{\textbf{Model}} &
  \multicolumn{3}{c}{\textbf{Rouge}} &
  \multicolumn{1}{l}{\multirow{2}{*}{\textbf{BLEU}}} &
  \multicolumn{1}{l}{\multirow{2}{*}{\textbf{BERTscore}}} &
  \multicolumn{1}{l}{\multirow{2}{*}{\textbf{Perplexity}}} \\ \cline{2-4}
               & Rouge1         & Rouge2         & RougeL         & \multicolumn{1}{l}{} & \multicolumn{1}{l}{} & \multicolumn{1}{l}{} \\ \hline
TinyLlama-1.1B & 0.267          & 0.191          & 0.239          & 0.09                 & 0.838                & 1.655                \\
Llama2-7B      & 0.469          & \textbf{0.363} & \textbf{0.434} & 0.11                 & 0.866                & \textbf{1.573}       \\
Mistral-7B     & \textbf{0.471} & 0.350          & 0.422          & \textbf{0.17}        & \textbf{0.875}       & 1.709                \\
BioMistral-7B  & 0.397          & 0.280          & 0.346          & 0.12                 & 0.865                & 1.970                \\
Meditron-7B    & 0.377          & 0.273          & 0.336          & 0.10                 & 0.853                & 1.592                \\
SOLAR-10.7B    & 0.442          & 0.313          & 0.386          & 0.12                 & 0.872                & 2.187                \\ \hline
\end{tabular}%
}
\caption{Performance of each fine-tuned model}
\end{table}
\begin{table}[]
\centering
\resizebox{\textwidth}{!}{%
\begin{tabular}{cccllll}
\hline
\textbf{Model} &
  \textbf{Accuracy} &
  \textbf{Completeness} &
  \textbf{R \& C} &
  \textbf{Consistency} &
  \textbf{Utility} &
  \textbf{Total} \\ \hline
Mistral-7B &
  4.4 &
  4.4 &
  \multicolumn{1}{c}{4.2} &
  \multicolumn{1}{c}{4} &
  \multicolumn{1}{c}{4.2} &
  \multicolumn{1}{c}{21.2} \\ \hline
\end{tabular}%
}
\caption{Qualitative assessment of generated discharge notes by Mistral-7B}
\end{table}
The qualitative assessment of the discharge notes generated by the Mistral-7B model is presented in Table 2. 
Five sample notes from the test set were evaluated by a cardiology expert across five criteria: Accuracy, Completeness, Readability and Comprehensibility, Consistency, and Utility, with each aspect rated on a 5-point scale.

Referring to the Table 2, one can observe the model's performance in generating clinically relevant and usable discharge documentation for patients. 
The scores reflect the model's capabilities in accurately capturing patient information, maintaining completeness of medical details, ensuring readability for the target audience, exhibiting consistent quality across samples, and providing documentation valuable for clinical decision-making and follow-up care planning.

While the specifics of the scores can be examined in Table 2, the overall assessment suggests the Mistral-7B model's proficiency in automated generation of discharge notes, meeting the standards expected in real-world healthcare settings.

\begin{table}[]
\centering
\resizebox{\textwidth}{!}{%
\begin{tabular}{lll}
\hline
\multicolumn{1}{c}{\textbf{Actual Discharge Note}} &
  \multicolumn{1}{c}{\textbf{Generated Discharge Note}} &
  \multicolumn{1}{c}{\textbf{Evaluation}} \\ \hline
\multicolumn{1}{l|}{\begin{tabular}[c]{@{}l@{}}{[}Chief Complaint{]}\\ nan\\ \\ {[}Operation and Procedure{]}\\ nan\\ \\ {[}Hospital Course{]}\\ ==========================\\ \#1. HTN\\ \#2. 1VD, pRCA total, small vessel disease \\     -\textgreater Medical Tx\\ \#3. AR\\ ==========================\\ \\ 0000.00.00 타원에서 시행한 coCT 상 RCA stenosis, \\ TTE 상 AR 확인됨.\\ \\ 0000.00.00 본원에서 F/E 원하여 CV OPD 내원함.\\ \\ 0000.00.00 CAG+ 위해 입원함.\\ \\ 0000.00.00 CAG+ 시행하였고 1VD, pRCA total 확인되나 \\                  small vessel disease 로 Medical Tx 하기로 함.\\                       TTE/TEE\textgreater @\\ \\ 0000.00.00 시술 후 급성 합병증 없이 퇴원함.\\ \\ {[}Condition at Discharge{]}\\ 호전됨\\ \\ {[}Type of Discharge{]}\\ 지시 퇴원\end{tabular}} &
  \multicolumn{1}{l|}{\begin{tabular}[c]{@{}l@{}}{[}Chief Complaint{]}\\ for CAG\\ \\ {[}Operation and Procedure{]}\\ nan\\ \\ {[}Hospital Course{]}\\ \#1. HTN\\ \#2. 1VD, pRCA total, small vessel disease \\     -\textgreater Medical Tx\\ \#3. AR\\ \\ \\ 상환 상병으로 본원 CV f/u 하시는 분으로\\ \\ 0000.00.00 CAG 시행 위해 입원함.\\ \\ 0000.00.00 CAG 시행하였고, pRCA total occlusion 확인되어 \\                   medical Tx 하기로 함.\\ \\ 0000.00.00 급성 합병증 없이 퇴원함.\\ \\ {[}Condition at Discharge{]}\\ 호전됨\\ \\ {[}Type of Discharge{]}\\ 지시 퇴원\end{tabular}} &
  \begin{tabular}[c]{@{}l@{}}1) Accuracy : 4/5\\ \\ \\ 2) Completeness : 5/5\\ \\ \\ 3) Readability \& Comprehensibility : 4/5\\ \\ \\ 4) Consistency : 4/5\\ \\ \\ 5) Utility : 4/5\\ \\ \\ Total : 21/25\end{tabular} \\ \hline
\multicolumn{1}{l|}{\begin{tabular}[c]{@{}l@{}}{[}Chief Complaint{]}\\ dyspnea\\ \\ {[}Operation and Procedure{]}\\ nan\\ \\ {[}Hospital Course{]}\\ ==========================\\ \\ \#1. 1VD \\      s/p PCI at p-mLAD (0000.00.00)\\ \\ \#2. DM\\ \\ \#3. HTN\\ \\ ==========================\\ \\ \textless{}\textless{}brief hx\textgreater{}\textgreater\\ \\ 상병 이외 특이 과거력 없는 환자로\\ \\ 0000.00.00 간헐적으로 발생하는 chest discomfort에 대해 \\                  CT cardiac stress perfusion, TTE 시행\\              CT 상 pLCX, dLCX stenosis 소견 확인\\ \\ 이후 opd f/u 하며 경과관찰\\ \\ 0000.00.00 증상 호전 없어 CAG+ 고려하 입원\\              pmLAD 1VD 확인되어 p-mLAD stent insertion 시행, \\                  diagonal branch POBA 시행 후 supportive care 위해 입원\\ \\ 0000.00.00 급성 합병증 없이 퇴원함.\\ \\ {[}Condition at Discharge{]}\\ 호전됨(improved)\\ \\ {[}Type of Discharge{]}\\ 지시 퇴원(planed discharge)\end{tabular}} &
  \multicolumn{1}{l|}{\begin{tabular}[c]{@{}l@{}}{[}Chief Complaint{]}\\ for PCI\\ \\ {[}Operation and Procedure{]}\\ nan\\ \\ {[}Hospital Course{]}\\ =======================\\ \#1. 1VD \\      s/p PCI at p-mLAD (0000.00.00)\\ \\ \#2. DM\\ \\ \#3. HTN\\ =======================\\ \\ 0000.00.00 건강검진 시행하였고, 검사결과 상 1VD 소견 보여\\ \\ 0000.00.00 상기소견에 대해 CAG 시행 위해 입원함.\\ \\ 0000.00.00 CAG 시행하였고, \\                  p-mLAD 에 90\% stenosis 소견 보여 PCI 시행함.\\ \\ 0000.00.00 급성합병증 없이 퇴원함.\\ \\ {[}Condition at Discharge{]}\\ 호전됨(improved)\\ \\ {[}Type of Discharge{]}\\ 지시 퇴원(planed discharge)\end{tabular}} &
  \begin{tabular}[c]{@{}l@{}}1) Accuracy : 5/5\\ \\ \\ 2) Completeness : 5/5\\ \\ \\ 3) Readability \& Comprehensibility : 5/5\\ \\ \\ 4) Consistency : 4/5\\ \\ \\ 5) Utility : 5/5\\ \\ \\ Total : 24/25\end{tabular} \\ \hline
\multicolumn{1}{l|}{\begin{tabular}[c]{@{}l@{}}{[}Chief Complaint{]}\\ for CAG\\ \\ {[}Operation and Procedure{]}\\ nan\\ \\ {[}Hospital Course{]}\\ 입원하여 CAG 시행하였으며 CAG 상 PCI 시행해야할 병변 있어 \\ 추후 staged PCI 시행하기로 하고 퇴원함.\\ \\ {[}Condition at Discharge{]}\\ 진단뿐\\ \\ {[}Type of Discharge{]}\\ 지시 퇴원\end{tabular}} &
  \multicolumn{1}{l|}{\begin{tabular}[c]{@{}l@{}}{[}Chief Complaint{]}\\ for CAG\\ \\ {[}Operation and Procedure{]}\\ nan\\ \\ {[}Hospital Course{]}\\ CAG 에서\\ \\ mRCA diffuse 50\% stenosis, FFR 0.91\\ pmLAD diffuse 50\% stenosis, FFR 0.91\\ \\ LAD PCI 시행하기로 함\\ \\ {[}Condition at Discharge{]}\\ 호전됨\\ \\ {[}Type of Discharge{]}\\ 지시 퇴원\end{tabular}} &
  \begin{tabular}[c]{@{}l@{}}1) Accuracy : 4 / 5\\ \\ \\ 2) Completeness : 4 / 5\\ \\ \\ 3) Readability \& Comprehensibility : 4 / 5\\ \\ \\ 4) Consistency : 4 / 5\\ \\ \\ 5) Utility : 4 / 5\\ \\ \\ Total : 20 / 25\end{tabular} \\ \hline
\end{tabular}%
}
\caption{Generated discharge notes by Mistral-7B and qualitative evaluation}
\end{table}
\section{Discussion}

The findings of this study emphasize the significant potential of fine-tuned LLM, particularly Mistral-7B, in automating the generation of discharge notes for cardiac patients. The quantitative results, evaluated across multiple metrics such as ROUGE, BLEU, BERT Score, and Perplexity, collectively demonstrate the models' ability to generate accurate, coherent, and clinically relevant discharge notes from patient medical records. While no single model excelled across all metrics, the overall performance highlights the promise of this approach in enhancing healthcare documentation efficiency and continuity of care.

Notably, the qualitative assessment by an expert in cardiology further reinforces the practical utility of the generated discharge notes. The Mistral-7B model exhibited a high degree of accuracy in capturing patients' medical conditions, treatment processes, and follow-up recommendations. The generated notes were deemed complete, including all essential medical information, while maintaining excellent readability and comprehensibility for the intended audience, including patients, guardians, and healthcare professionals. Additionally, the consistent quality across various patient records and the overall utility in supporting clinical decision-making and care planning underscore the model's potential for real-world implementation in healthcare settings.

\textbf{Limitations} While the quantitative metrics employed, such as ROUGE, BLEU, BERT Score, and Perplexity, provide valuable insights into the models' performance, they are not specifically designed to evaluate medical documentation. These metrics may not fully capture the nuances and critical aspects of healthcare documentation, such as adherence to clinical guidelines, appropriate use of medical terminology, and patient safety considerations. The lack of domain-specific, clinically-oriented evaluation metrics poses a challenge in comprehensively assessing the generated discharge notes' quality and suitability for healthcare applications. 
Additionally, the lack of standardization in the formatting and style of progress notes poses a challenge for model training and performance. Progress notes are often written in a free-form manner by different physicians, leading to inconsistencies in aspects such as date formatting, abbreviations, and terminology usage. For instance, some physicians may write dates as "210304," while others use formats like "2021/03/04." Such variations can introduce confusion and inconsistencies during the model's learning process, potentially impacting its ability to accurately interpret and generate discharge notes. Establishing guidelines or implementing preprocessing steps to standardize the input data could mitigate these issues and improve the model's performance.

\textbf{Future works} Firstly, the development and integration of domain-specific, clinically-oriented evaluation metrics tailored for assessing medical documentation is crucial. These metrics should capture essential aspects such as adherence to clinical guidelines, appropriate use of medical terminology, and patient safety considerations, enabling a more comprehensive and reliable evaluation of the generated discharge notes.

Another crucial direction for future research is the extension of this approach to other medical specialties beyond cardiology. While this study focused on generating discharge notes for cardiac patients, the methodologies and insights gained could be adapted and applied to other domains, such as oncology, neurology, or pediatrics. By expanding the scope of the dataset to encompass diverse medical conditions and specialties, the models could be fine-tuned to generate discharge notes tailored to the specific requirements and terminologies of each field. This would not only broaden the applicability of the AI-driven documentation approach but also contribute to a more comprehensive and unified system for streamlining medical documentation across various healthcare domains.

Furthermore, expanding the dataset to include data from multiple healthcare facilities and diverse patient populations could significantly enhance the models' generalizability and robustness, ensuring their applicability across different clinical settings and patient demographics. This could involve establishing collaborations with other healthcare institutions or leveraging existing large-scale medical datasets.

The exploration of multi-modal approaches, incorporating not only textual data but also medical images, lab results, and other relevant patient information, presents a promising direction. By leveraging these diverse data sources, the models could gain a more holistic understanding of patients' conditions, enabling the generation of more accurate and well-rounded discharge notes. Additionally, the integration of advanced techniques such as attention mechanisms and multimodal fusion could further improve the models' ability to effectively utilize and combine information from multiple modalities.

\textbf{Conclusion} This study represents a significant step towards leveraging the power of large language models in revolutionizing healthcare documentation practices. The promising results demonstrate the potential for automating discharge note generation, reducing administrative burdens on healthcare professionals, and facilitating seamless continuity of care for patients. However, continued research and development, including the incorporation of domain-specific evaluation metrics and the exploration of multi-modal and knowledge-augmented approaches, are crucial for further refining and advancing these AI-driven solutions in the healthcare domain.

\bibliography{colm2024_conference}
\bibliographystyle{colm2024_conference}

\end{document}